\documentclass{article}
\usepackage{spconf,amsmath,graphicx}
\usepackage{interval}
\usepackage{todonotes}

\usepackage{floatrow}
\newfloatcommand{capbtabbox}{table}[][\FBwidth]
\usepackage{adjustbox}
\usepackage{amsfonts}
\usepackage{balance}

\usepackage{pict2e}

\def\L{{\cal L}}
\title{Joint Coarse-and-Fine Reasoning for Deep Optical Flow}

\name{Victor Vaquero$^{\star}$ \enskip German Ros$^{ \mathsection}$  \enskip Francesc Moreno-Noguer$^{\star}$ \enskip Antonio M. Lopez$^{\dagger \wr}$ \enskip Alberto Sanfeliu$^{\star}$}

\address{$^{\star}$ Institut de Rob\`otica i Inform\`atica Industrial, CSIC-UPC, Barcelona, Spain \\
    $^{\wr}$ Universitat Aut\`onoma de Barcelona, Campus UAB, Barcelona, Spain\\
    $^{\dagger}$ Computer Vision Center, Campus UAB, Barcelona, Spain\\
    $^{\mathsection}$ Toyota Research Institute, Palo Alto, USA}

\begin{document}
%\ninept
%
\maketitle
\begin{abstract}
We propose a novel representation for dense pixel-wise estimation tasks using CNNs that boosts accuracy and reduces training time, by explicitly exploiting joint coarse-and-fine reasoning. The coarse reasoning is performed over a discrete classification space to obtain a general rough solution, while the fine details of the solution are obtained over a continuous regression space. In our approach both components are jointly estimated, which proved to be beneficial for improving estimation accuracy. Additionally, we propose a new  network architecture, which combines coarse and fine components by treating the fine estimation as a refinement built on top of the coarse solution, and therefore adding details to the general prediction. We apply our approach to the challenging problem of optical flow estimation and empirically validate it against state-of-the-art CNN-based solutions trained from scratch and tested on large optical flow datasets.
\end{abstract}

%

%%												%%
% 			Victor Vaquero   -  ICIP 2017		 %
% 			   									 %
% 				www.victorvaquero.me			 %
%%												%%

\section{INTRODUCTION}
\label{sec:intro}

Deep Convolutional Neural Networks (CNNs) have become a de facto standard
to successfully address all kind of perception related problems, such as image classification, object detection and optical flow. Fresh CNN architectures and training procedures are day after day becoming the new state of the art, producing models which prediction accuracy was inconceivable few years ago. The ascendancy of these tools is greatly due to the release of very large annotated datasets as well as the popularization of massively-parallel GPUs, which enable fast training and inference.

In addition to the two aforementioned elements, the success of CNN-based approaches heavy rely on a smart design of three elements, which are i) the representation of the problem, ii) the training method and iii) the network architecture.

\begin{figure}[t]
\centering
  \centerline{
  	\includegraphics[width=0.99\textwidth]{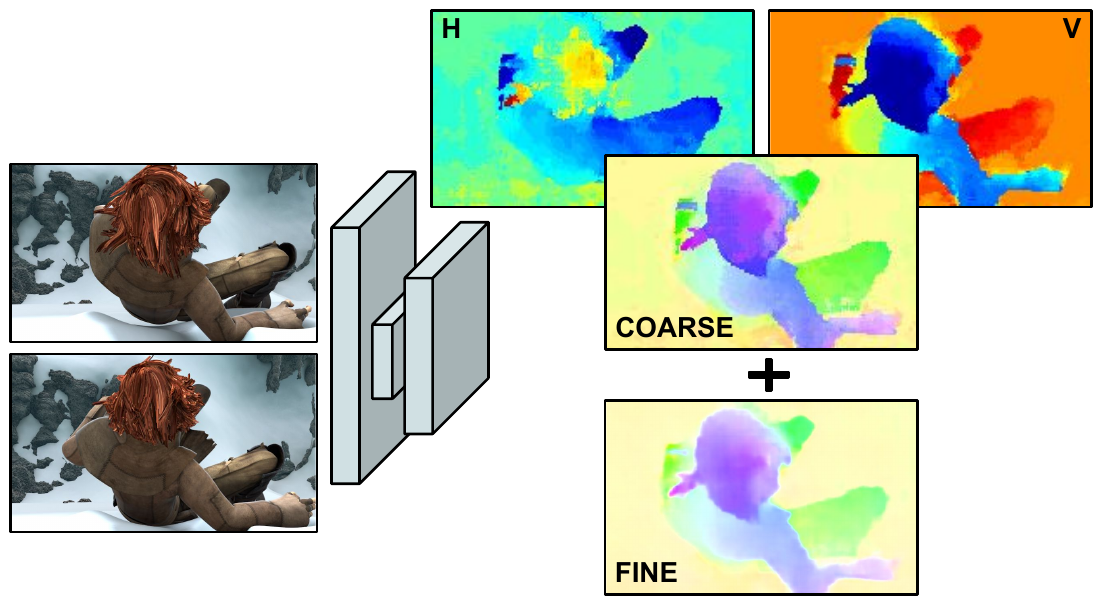}}
\vspace{-2mm}
  
  \caption{We approach dense per-pixel regression problems with a joint coarse-and-fine method and apply it to the challenging problem of optical flow prediction. %, boost the performance of state-of-the-art CNNs regression-only networks trained from scratch. 
  Our method explicitly combines a coarse result based on the solution of pixel-wise horizontal and vertical classification problems with a fine one obtained through regression predictions.}
  
  \label{fig:general}
  \vspace{-4mm}
\end{figure}

As a general practice, the task under study is represented as a set of classification or regression problems, depending on the nature of the task. For example, it is common to represent semantic segmentation~\cite{RosArxiv16} as multiple classification problems over a finite and discrete set of categories, while motion related tasks ---such as optical flow prediction~\cite{dosovitskiy2015flownet}--- are represented as regression problems over a continuous ''flow space''. To promote the correct behavior of CNNs, the training method needs to reflect the chosen representation with an appropriate loss function. Typical examples of losses are cross entropy and mean squared error, associated to classification and regression problems respectively. The last element, the network architecture, needs to provide enough capacity for the approximation of the task and support the propagation of the gradient to make the training possible. The use of certain network designs, as for instance the architectures based on residual blocks has proven to yield a notorious improvement in speed and accuracy~\cite{He_2016_CVPR}.

Unlike existing approaches, in this paper we propose an alternative representation that combines the benefits of classification and regression in a joint coarse-and-fine reasoning as shown in Fig.~\ref{fig:general}. 
The classification component carries general coarse information that is important to focus the search around the solution space, while the regression component carries the fine details needed to produce an accurate prediction. 
We defend that this representation is more suitable that the existing ones, helping to reach better solutions faster. To enforce this joint representation we propose one simple but effective loss functions that linearly combines a classification and a regression cost. 
We also show how to fully integrate this representation in any network architecture by introducing a new layer that expresses the final prediction as the addition of a refinement real component on top of a coarse discrete approximation. 

Our approach is applied to the context of optical flow due to its challenging nature, where a real value needs to be predicted for each pixel of an image that may follow any kind of motion. We demonstrate the benefits of our proposal in state-of-the-art optical flow datasets.

% -------------------------------------------------------------------------------------------- % 
% -------------------------------------------------------------------------------------------- % 

\section{RELATED WORK}
\label{sec:soa}

The formulation of optical flow approaches has continued evolving from the classical energy optimization formulation over a pixel-brightness space~\cite{horn1981determining} to sophisticated variational approaches~\cite{sun2014quantitative}, which include all type of ad hoc blocks to account for key aspects such as edges motion~\cite{revaud2015epicflow} and robust patch matching~\cite{Bailer_2015FlowFields}. This evolution towards improving flow accuracy brought the addition of object semantics~\cite{sevilla2016optical} and eventually the will of exploiting semantic information and context to improve flow estimation led to approach the task as a learning problem, exploiting the power of CNN-based techniques. 

It is clear that CNNs~\cite{lecun1998gradient}\cite{krizhevsky2012imagenet} have gained much attention in the context of optical flow. They have been applied to improve many different parts of the pipeline, from dealing with image patch matching in large motion displacement~\cite{weinzaepfel2013deepflow}\cite{revaud2015epicflow}, to the extraction and match of features patch~\cite{bai2016exploiting}. The first totally CNN-based optical flow approach was introduced in~\cite{dosovitskiy2015flownet}, where authors show that it is feasible to reach state-of-the-art solutions training a CNN architecture end-to-end. Such an approach builds upon the recent success of deconvolutional blocks to solve dense pixel-wise prediction problems, such as semantic segmentation~\cite{zeiler2010deconvolutional}\cite{Long_2015_CVPR}\cite{noh2015learning}\cite{RosArxiv16} and super resolution~\cite{dong2016image}. Most of the CNN-based solutions typically address the learning task casting it to a classification or a regression problem, depending on the nature of the task. In our novel approach, we perform joint classification and regression to exploit their respective benefits, i.e., i) obtaining a simplified coarse solution via classification, which helps the training to converge quicker and ii) distilling the fine details of a solution via regression. We prove that this approach leads to better results than existing coarse-to-fine strategies used in methods like FlowNet~\cite{dosovitskiy2015flownet}, where the problem is hierarchically approached from low to high resolution.

%%												%%
% 			Victor Vaquero   -  ICIP 2017		 %
% 			   									 %
% 				www.victorvaquero.me			 %
%%												%%

\section{THE COARSE-AND-FINE FORMULATION}
\label{sec:typestyle}

Optical flow, as well as many other dense pixel-wise prediction tasks, is traditionally formulated as a regression problem in order to predict a solution that is intended to capture fine details. However, in this work we defend that it is more convenient and accurate to jointly represent a coarse classification component, which contains a generic and discrete approximation to the solution, and a fine regression component, which provides a fine and continuous refinement. 
The introduction of an explicit discrete classification term draws inspiration from semantic segmentation methods, which exhibit fast convergence rates. In our case, this component helps on accelerating the training by quickly centring the search space around a coarsely correct solution.

Here we describe the concepts and ingredients used to fully exploit this joint coarse-and-fine representation, including two different network topologies with their respective training methods and associated loss functions.

\begin{figure*}[t]
\centering
 \centerline{
 	\includegraphics[width=1\textwidth]{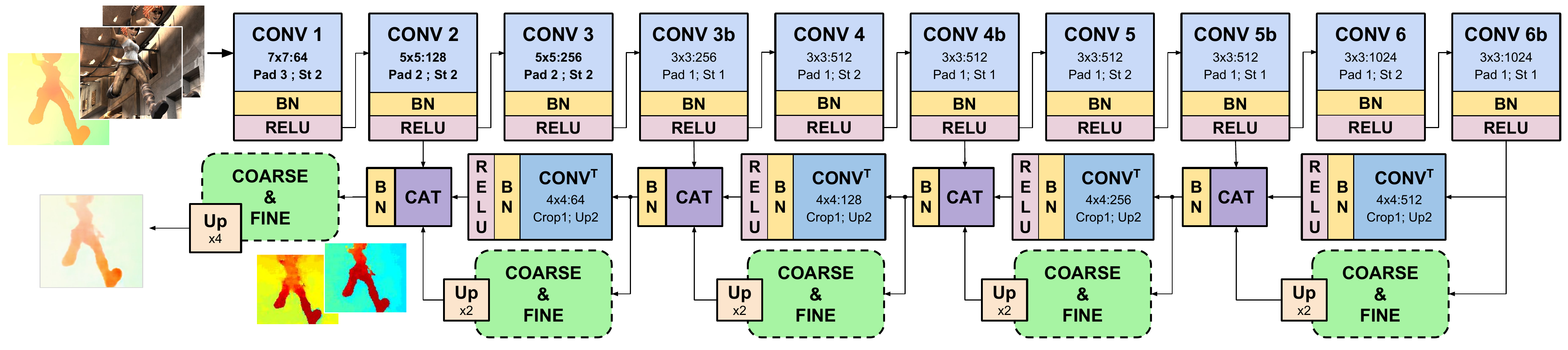}}
\vspace{-3mm}
 \caption{Our regularised FlowNet architecture is formed of a contractive and an expansive part. First, a set of Convolutions, Batch Normalization and Relu layers (CONV, BN and RELU) are interleaved to obtain abstract and hierarchical representations while contracting the input information. The final dense predictions is generated by deconvolution layers (CONV$^T$), and guided to an optimal solution by concatenating (CAT blocks) corresponding initial feature maps and the partial coarse and fine upsampled (Up) solutions obtained at five resolution points.}
 
 \label{fig:caf_diagram}
 \vspace{-3mm}
\end{figure*}

\subsection{Estimating coarse information as an auxiliary task}\vspace{-2mm}

For the sake of generality, we define a basic architecture for the estimation of optical flow as a combination of two blocks, $F_\theta(\cdot)$ and $G_\theta(\cdot)$. Given an RGB image $\mathcal{X} \in \mathbb{R}^{H \times W \times 3}$, an initial stage of the network computes features $F_\theta(\mathcal{X})$ according to the network model $\theta$. Then, a second stage transforms these features into pixelwise optical flow predictions $G_\theta(F_\theta(\mathcal{X})) = \hat{\mathcal{Y}} \in \mathcal{R}^{H \times W \times 2}$. In our approach, we adapt FlowNet~\cite{dosovitskiy2015flownet} to use it as our $F_\theta$ (see Fig.~\ref{fig:caf_diagram} for a graphical description).

Traditional CNN-based methods define $G_\theta$ as a set of convolutions that transform the extracted representation into the final optical flow. Then, during training a regression loss function is used to find a suitable model $\theta$, and therefore just using fine-grained information.

A simple way to account for both coarse-and-fine components is to branch $G_\theta$ into $G_\textit{class} =\hat{\mathcal{Y}}^\textit{class}$ and $G_\textit{reg} = \hat{\mathcal{Y}}^\textit{reg}$. Here, $\hat{\mathcal{Y}}^\textit{reg}$ stands for a fine-grained regression  solution and $\hat{\mathcal{Y}}^\textit{class}$ stands for a coarse classification prediction over a small set of flow categories. 

$G_\textit{reg}$ is given by a $3 \times 3$ convolution kernel, which maps to a 2 channel output representing flow. On the other hand, $G_\textit{class}$ consists of a simple $3\times 3$ convolution mapping to $K$ categories and followed by a soft-max operator. These K categories are defined by projecting  optical flow within the $[m_r, M_r]$ range, which bounds are empirically selected according to typical minimum and maximum values for this problem. Then this range is divided into $K$ categories $I_k$ such that: 

\vspace{-5mm}
\begin{equation}%
I_k =
\left\{%
	\begin{array}{ll}
		(-\inf, C_1+\delta/2), \text{ if } k = 1 \\
		\left[C_k-\delta/2,C_k+\delta/2\right) \text{ if } 1 < k < K \\
		\left[C_K - \delta/2, +\inf \right) \text{ if } k=K
	\end{array}
\right\}.
\end{equation}

 $C_k = m_r + \delta(k-1) , k \in {1,\dots,K}$ are the centroids of the classes. Notice that outbound pixels are codified on the outer classes. This procedure serves also to transform the regression ground truth ${\mathcal{Y}}^\textit{reg}$ into classification ground truth ${\mathcal{Y}}^\textit{class}$.

During training $\theta$ is adjusted via standard end-to-end back-propagation, guided by the following  coarse-and-fine loss function:
\vspace{-0mm}
\begin{equation}%
%\vspace{-2mm}
\label{eq:CaF}
	\L_{CaF}(\hat{\mathcal{Y}}, \mathcal{Y}) =
	\L_\textit{coarse}(\hat{\mathcal{Y}}^\textit{ class}, {\mathcal{Y}}^\textit{class}) + \lambda \L_\textit{fine}(\hat{\mathcal{Y}}^\textit{ reg}, {\mathcal{Y}}^\textit{reg}).
\end{equation}

In our approach, $\L_\textit{fine}$ is a standard $\ell_2$-norm. For  $\L_\textit{coarse}$, we use the multi-class Weighted Cross Entropy loss (WCE)~\cite{RosArxiv16}, such that:

\vspace{-5mm}
\begin{equation}
\L_{WCE} = -\sum_{i,j,k}^{H,W,K} \omega(\mathcal{Y}^\textit{ class}_{i,j}) \textit{Id}_{[\mathcal{Y}^\textit{ class}_{i,j}]} (\text{log}(\hat{\mathcal{Y}}^\textit{ class}_{i,j,k})), 
\end{equation}

\noindent where $\textit{Id}_{[x']}(x)$ is an index function that acts as a selector for the probability associated to the expected ground truth class. $\omega(k)$ is a weight proportional to the inverse of the frequency of the $k$-th class, which is a key factor to prevent the bias introduced by class imbalance due to some predominant vector flows and is computed from the training set statistics.

We refer to this approach as CaF. In practice, and without loss of generality, we further sub-divide $G_\textit{class}$ into its horizontal and vertical optical flow terms $\hat{\mathcal{Y}}^\textit{class-H}$, $\hat{\mathcal{Y}}^\textit{class-V}$ to simplify the representation of the problem. In the CaF, coarse and fine components are never combined. The output of the network is just the regression, while the coarse component is used as an an auxiliary task to provide additional guidance and speed up to the training process. 
Despite its simplicity, this method serves us to test and validate the importance of accounting for both coarse and fine components.

\subsection{Explicit Joint Coarse-and-Fine}\vspace{-2mm}
\label{sec:cafArch}

\begin{figure*}[t]
\vspace{-1mm}

\begin{floatrow}
\ffigbox[\FBwidth]{%
      \includegraphics[width=0.355\textwidth]{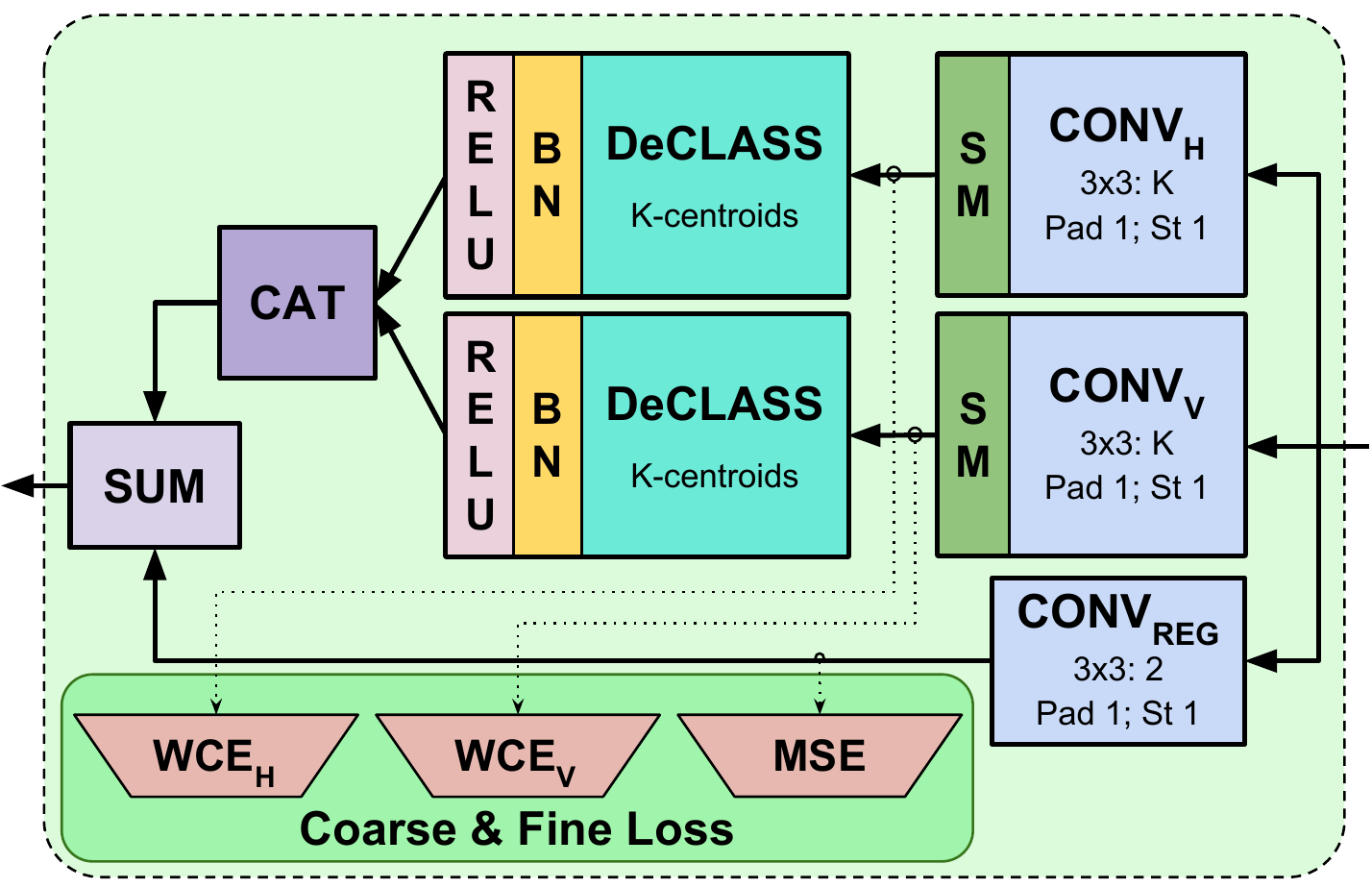}
}{%
\vspace{-5.5mm}
  \caption{
  Each coarse-and-fine module solves one regression and two classification per-pixel problems. Softmax outputs from these last are declassified obtaining the coarse solution.}
\label{fig:caf}
}

\capbtabbox{%
\begin{adjustbox}{max width=0.58\textwidth}
\begin{tabular}{l||c|c|c|c|c|}%\hline
\multicolumn{1}{l||}{} & \multicolumn{1}{|c|}{F.Chairs}&\multicolumn{2}{|c|}{{Sintel Train}} &\multicolumn{2}{|c|}{{Sintel Test}}\\

{ } &\multicolumn{1}{|c|}{Validation}&\multicolumn{1}{|c|}{Clean}&\multicolumn{1}{|c|}{Final} &\multicolumn{1}{|c|}{Clean}&\multicolumn{1}{|c|}{Final}\\

\hline
\hline %& {-}& {-}& {-}
%\multicolumn{15}{c}{\em {Single Object}}\\
{Regression}&  {3.78 (100)}     & 	{6.93 (100)} & {7.66 (100)}   & {9.98 (100)}     & {10.72 (100)}\\
\hline

{Class-5c}   & {6.99 (184.7)}    & {9.66 (139.4)} & {10.20 (133.1)}&	{13.11 (131.3)} & {13.54 (126.3)}\\
{Class-21c}  & {4.06 (107.3)}    & {7.91 (114.1)} & {8.50 (110.9)} & {10.70 (107.1)} & {11.34 (105.8)}\\
{Class-41c}  & {3.81 (100.7)}    & {7.69 (110.87)}& {8.38 (109.3)} & {10.66 (106.7)} & {11.53 (107.5)}\\
\hline

{CaF-5c}    &  {3.55 (93.8)}    & {6.85 (98.8)} & {7.54 (98.5)}   & {9.98 (99.9)}    & {10.69 (99.7)}\\
{CaF-21c}   &  {3.44 (90.9)}    & {6.76 (97.5)} & {7.43 (96.9)}   & {9.88 (98.9)}    & {10.53 (98.2)}\\
{CaF-41c}   &  {3.47 (91.7)}    & {6.75 (97.4)} & {7.39 (96.4)}   & {9.77 (97.8)}    & {10.48 (97.7)}\\
\hline

{CaF-Full-5c} & {3.25 (85.8)}    & {6.85 (98.84)} & {7.72 (100.7)} &	 {9.74 (97.5)}  & {10.51 (98.1)}\\
{CaF-Full-21c}& {3.23 (85.3)}    & {6.75 (97.34)} & {7.59 (99.0)}  &	 {9.57 (95.8)}  & {10.28 (95.9)}\\
{CaF-Full-41c}& {3.18 (84.0)}    & {6.51 (93.84)} & {7.28 (95.0)}  &	 {9.42 (94.3)}  & {10.18 (95.0)}\\
\hline
\hline

\end{tabular}
\end{adjustbox}
}{%
\vspace{-5mm}
    \caption{Evaluation of the end-point-error for the presented models. Suffixes $K$c indicate the number of classes used during training on Flying Chairs. Results over the Sintel Training and Test sets are presented, showing the generalization of our method on unseen datasets.}
    \label{tab:results}
}
\end{floatrow}
\vspace{-2mm}
\end{figure*}

We propose a refinement of the previous approach that explicitly represents the optical flow estimation by adding the output of the regressor to the classifier component. In this case, the regressor does not encode the whole optical flow, but just the fine details of the solution, i.e., a refinement, which is combined with the coarse solution provided by the horizontal and vertical classifiers to produce the final estimation. This process, that we call CaF-Full, is depicted by Fig.~\ref{fig:caf}. This representation has the advantage of reducing the search space of the fine component to a bounded area around zero, which makes the training convergence faster and leads to more accurate models (see section~\ref{sec:exps}).

In practice, the combination of the three components, i.e., $\hat{\mathcal{Y}}^\textit{class-H}$, $\hat{\mathcal{Y}}^\textit{class-V}$ and $\hat{\mathcal{Y}}^\textit{res}$ requires to map the discrete classification solutions back to a real value. This is done in the DeCLASS blocks (Fig.~\ref{fig:caf}), which output the centroid associated to a given class. Afterwards horizontal and vertical components are concatenated and added to the regression output.

%%												%%
% 			Victor Vaquero   -  ICIP 2017		 %
% 			   									 %
% 				www.victorvaquero.me			 %
%%												%%

\section{EXPERIMENTS}
\label{sec:exps}

In our experiments, we take a state-of-the-art regression-based CNN architecture~\cite{dosovitskiy2015flownet} and validate the benefits of adding our joint coarse-and-fine reasoning scheme in terms of optical flow end-point-error (EPE). As additional baselines classification-only and regression-only predictions are also reported. Experiments are summarized in Table~\ref{tab:results}, where we show that our proposal decreases EPE by up to a $15\%$.

\subsection{Experimental conditions}\vspace{-2mm}

All the presented models are trained from scratch under the exact same conditions, allowing to measure the real performance boost that our approach produce.
FlyingChairs \cite{dosovitskiy2015flownet} is used for training, adopting the same splits than the original paper and a batch size of $8$ pairs of images as input. 
We perform slight data augmentation by mirroring upside-down and left-to-right the images with a $50\%$ chance each. 
All models are implemented in MatConvNet, initialized following He's method~\cite{he2015delving} and trained using Adam with the standard parameters $\beta_1 = 0.9$ and $\beta_2 = 0.999$.
The training process is performed on a single NVIDIA K40 GPU for $600,000$ epochs, fixing the learning rate to $10^{-3}$ during the first $300,000$ epochs and successively halving it each $100,000$ epochs. 
Following~\cite{dosovitskiy2015flownet} we measure the network loss at $5$ different resolution points on the expansive part (Fig.\ref{fig:caf_diagram}), but contrary to their approach we weight all these losses equally to avoid extreme hyper-parameter tuning. 
For the coarse prediction, we bound the continuous flow space between $-40$ and $40$ (parameters $m_r$ and $M_r$ respectively), and discretise the resulting subdomain. We perform three different experiments attending to the number of classes created and therefore the size of the pixel flow bins ($\delta$). We choose to test $5$, $21$ and $41$ classes, each one representing flow ranges of $20$, $10$ and $2$ respectively. 

\vspace{-2mm}
\subsection{Regression baseline}\vspace{-2mm}
Our regression baseline consists of a Batch-Normalized FlowNet trained from scratch under the previously defined conditions.
The regression baseline is trained by deactivating the contribution of the classification modules to the final output as well to the loss function (turning off the upper part of Fig.~\ref{fig:caf}).
The reported results are fairly close to the ones of the original paper, but we used moderated data augmentation and avoided the hyper-parameter tuning in order to create a fair and reproducible test environment. Notwithstanding, the increase in performance of our joint approach is evident, as the training procedure is rigorous and fixed for all the methods. 

\vspace{-2mm}
\subsection{Classification baseline}\vspace{-2mm}
In addition to the regression baseline, Table~\ref{tab:results} reports classification results labelled as Class-Kc, for $K = \{5, 21, 41\}$ classes. 
This baseline is trained by deactivating the regression contribution to the network output (the "SUM" block in Fig.~\ref{fig:caf}) as well as the MSE error of the loss during training, so that only the coarse components are used.

\vspace{-2mm}
\subsection{Joint Coarse-and-Fine performance}\vspace{-2mm}

We report experiments for the two flavours of our proposal, i.e., i) \textit{CaF}, which is the regression baseline trained with the proposed coarse-and-fine loss function ---turning the DeClass modules of Fig.~\ref{fig:caf} off, but keeping its measured errors on---, and ii) our full coarse-and-fine proposal (\textit{CaF-Full}) where the coarse-and-fine refinement is plugged, explicitly creating the network output in that way.

According to the results, the performance boost produced by our approach in the trained networks is significant. The addition of the combined loss function (see Table~\ref{tab:results} rows 5--7) noticeably decreases the end-point-error (EPE). 
Moreover, by introducing our full coarse-and-fine architecture (rows 8--10), described in section~\ref{sec:cafArch}, the performance is boosted up to a $15\%$ in the Flying chairs validation set.

Regarding the number of classes of the coarse prediction, we observe a trend in the full architecture as the error tends to decrease with the number of classes. This is more clear for the \textit{CaF-Full} models, has having smaller class bins allows the fine prediction to recover misclassified pixels easier.

We further evaluate the generalization capacities of our approach by testing the models trained on FlyingChairs over the unseen Sintel dataset without any finetuning. Although the improvement is not so abrupt in this challenging dataset, the same conclusions can be systematically obtained for both training and test Sintel splits. This  proves once more the benefits of our joint Coarse-and-Fine methods.

\vspace{-1mm}
\section{CONCLUSIONS AND FUTURE WORK}
\label{sec:conc}
\vspace{-3mm}
{}\footnote{Acknowledgements: This work was partially supported by European AEROARMS project (H2020-ICT-2014-1-644271) and CICYT projects ColRobTransp (DPI2016-78957-R), ROBINSTRUCT (TIN2014-58178-R). Authors thank Nvidia for GPU hardware donation.}This paper presented the benefits of using a joint coarse-and-fine representation for dense pixel-wise estimation task---such as optical flow--- by casting the task to a joint classification and regression problem. Our novel representation has proven to speed up training convergence and to increase model accuracy when compared against CNN-based state-of-the-art methods and other baselines. We have experimentally demonstrated that this joint representation achieves its maximum potential by exploiting a new type of architecture, which expresses its prediction as the addition of a refinement real component to a coarse discrete approximation. Our next steps are focused on the study the impact that complementary sources of information have in models accuracy and how to efficiently combine those sources.

\balance

\bibliographystyle{IEEEbib}
\bibliography{refs}

\end{document}